%% file: TTRL_VAU_arXiv.tex
\documentclass[letterpaper]{article} % DO NOT CHANGE THIS
\usepackage[preprint]{2027}  % Show authors and suppress the proceedings copyright notice for arXiv.
\usepackage[hyphens]{url}  % DO NOT CHANGE THIS
\usepackage{graphicx} % DO NOT CHANGE THIS
\usepackage{natbib}  % DO NOT CHANGE THIS AND DO NOT ADD ANY OPTIONS TO IT
\usepackage{caption} % DO NOT CHANGE THIS AND DO NOT ADD ANY OPTIONS TO IT
\usepackage{algorithm}
\usepackage{algorithmic}
\usepackage{amsmath}
\usepackage{multirow}
\usepackage{amssymb}
\usepackage{newfloat}
\usepackage{listings}
\DeclareCaptionStyle{ruled}{labelfont=normalfont,labelsep=colon,strut=off} % DO NOT CHANGE THIS
\floatstyle{ruled}
\newfloat{listing}{tb}{lst}{}
\floatname{listing}{Listing}

\usepackage{booktabs}

\usepackage{multirow}
\usepackage[table]{xcolor}
\usepackage{booktabs}
\usepackage{subcaption}
\definecolor{oursblue}{RGB}{232,244,252}

\title{Test-time Reinforcement Learning for Anomalous Video Understanding}
\author{
    Huining Li\textsuperscript{\rm 1},
    Yuxiang Duan\textsuperscript{\rm 1},
    Jiyang Tan\textsuperscript{\rm 1},
    Qian Li\textsuperscript{\rm 1},
    MingCai Chen\textsuperscript{\rm 2},
    Jian Zhang\textsuperscript{\rm 3},\\
    Xingdong Sheng\textsuperscript{\rm 3}\corresponding,
    Yuntao Du\textsuperscript{\rm 1}\corresponding
}
\affiliations{
    \textsuperscript{\rm 1} C-FAIR \& School of software, Shandong University, 
    \textsuperscript{\rm 2} Nanjing University of Posts and Telecommunications\\
    \textsuperscript{\rm 3} Lenovo Research\\
}

\begin{document}

\maketitle

\begin{abstract}
Anomalous video understanding aims to identify abnormal events in videos and interpret their semantic meanings beyond simple anomaly detection. 
Recent video large language models (Video-LLMs) have demonstrated promising zero-shot capabilities for this task, yet their performance remains limited due to insufficient adaptation to diverse anomaly patterns and evolving environments. 
Test-time reinforcement learning offers a promising solution by enabling models to improve through self-generated feedback signals without requiring additional human annotations.  
However, applying it to anomalous video understanding remains challenging due to three issues: (1) generated pseudo-labels can be unreliable when consensus is weak; (2) binary reward designs fail to capture uncertainty in model generations, resulting in ineffective optimization signals; and (3) unanimous rollout groups receive identical rewards, causing group-relative advantages to collapse and eliminating effective policy-gradient signals.
To address these challenges, we present a novel test-time reinforcement learning framework for anomalous video understanding by introducing dual-query consistency filtering, an entropy-aware consensus reward, and a virtual negative anchor mechanism. The framework retains reliable samples through consistency across semantically equivalent queries, combines answer agreement with generation uncertainty for reward estimation, and introduces a virtual negative anchor to create reward variation in unanimous rollout groups, thereby preserving effective group-relative optimization signals. 
Experiments on VAU-Bench show that our method outperforms the compared frozen and supervised baselines. The gains are most pronounced on the ECVA subset of VAU-Bench with thinking, where accuracy improves from 75.81\% to 90.00\% relative to the frozen backbone.
% Our code is available at \url{https://github.com/lihuining617/TTRL-AVU}.
\end{abstract}

\input{section/introduction}
\input{section/related_work}
\input{section/method}

\input{section/experiment}

\section{Conclusion}

In this work, we presented a test-time reinforcement learning framework for anomalous
video understanding, enabling a Video-LLM to adapt
continuously from unlabeled test videos rather than remaining frozen
after offline training. The proposed framework addresses three key
limitations of directly applying existing test-time reinforcement learning methods to this task. Specifically, dual-query consistency
filtering reduces the influence of unreliable pseudo-supervision, the
entropy-aware consensus reward provides a more informative learning
signal than binary majority rewards, and the virtual negative anchor prevents
group-relative advantages from vanishing for unanimous rollout groups.
These components are integrated into a GRPO objective for
stable online adaptation throughout the incoming unlabeled test stream.
Experiments on VAU-Bench demonstrate that the proposed method improves
anomalous video understanding under both the w/o Think and w/ Think prompting settings.
The adapted models also exhibit useful transfer to anomaly classification without additional
classification-specific parameter updates, with consistent improvements over the frozen
Qwen2.5-VL-3B backbone across all VAU-Bench subsets.
Future work will investigate more efficient adaptation strategies and stronger
mechanisms for preserving fine-grained class boundaries during
continuous test-time learning.

\newpage
\bibliography{2027}


\end{document}

%% file: section/introduction.tex
\section{Introduction}

\begin{figure}[t]
\centering
\includegraphics[width=\columnwidth]
{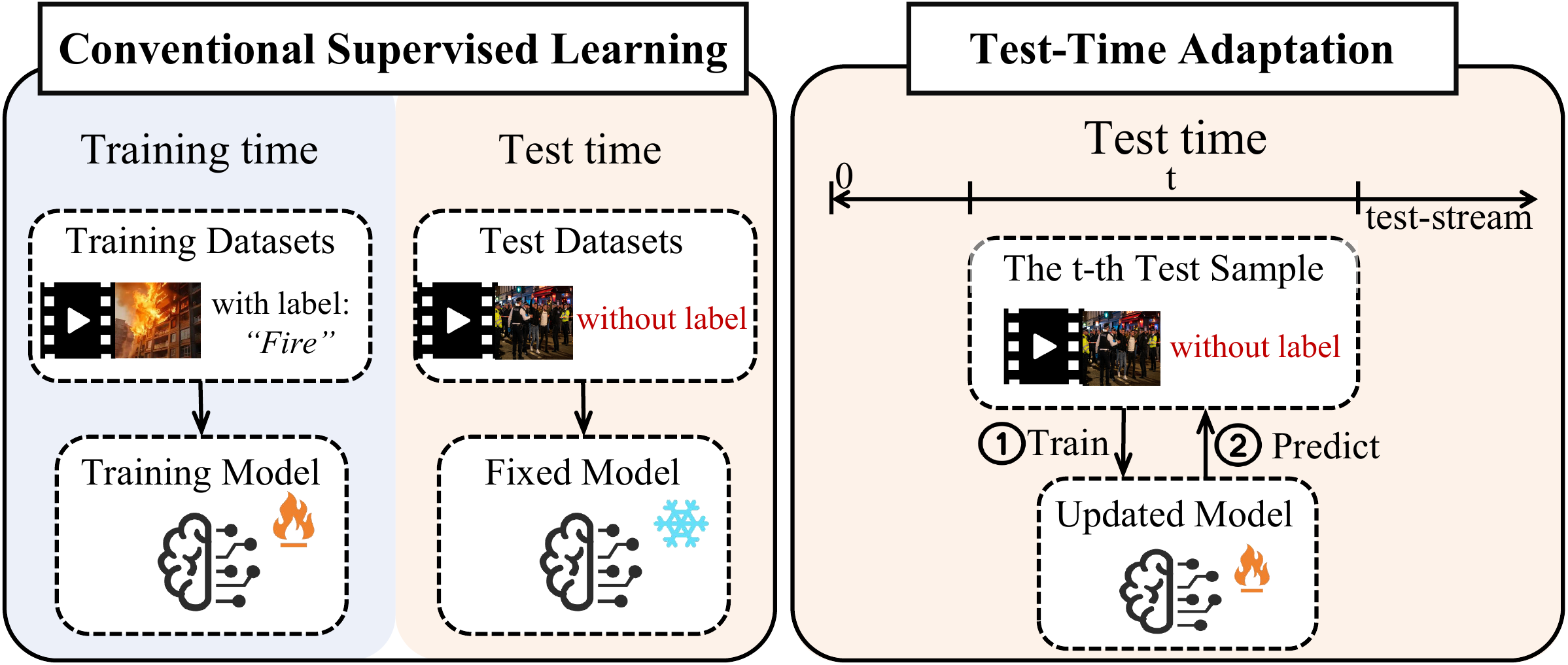}
\caption{
Comparison between conventional supervised learning and test-time adaptation.}
\label{fig:setting_comparison}
\vspace{-3mm}
\end{figure}

Anomalous video understanding~\cite{ye2025vera} aims to detect and semantically interpret abnormal events by identifying anomalies and understanding the involved objects, actions, interactions, and event semantics. This capability is important for public surveillance, industrial monitoring, and emergency response, where accurate interpretation enables timely intervention and informed decision-making.

Recent Video-LLMs have shown promising zero-shot performance on this task~\citep{lin2024videollava,zhang2025llavavideo}. However, they remain limited in complex scenarios with diverse anomaly patterns and evolving deployment environments. Most existing methods rely on offline training and keep model parameters frozen during inference, limiting their ability to handle distribution shifts in scenes, viewpoints, environments, and anomaly types~\cite{sun2020ttt,wang2022cotta}. Repeated retraining is also impractical because anomalous events are rare and fine-grained annotation is costly~\citep{sultani2018ucfcrime,holmesvau}.

Fortunately, test-time adaptation~\cite{sun2020ttt,wang2021tent,wang2022cotta,niu2022eata} offers a promising alternative by learning directly from incoming test samples during inference without ground-truth labels. As illustrated in Figure~\ref{fig:setting_comparison}, the model is updated on each unlabeled sample before prediction, with the adapted parameters carried forward to subsequent samples. This paradigm exploits the test stream without human annotation and has been extended to vision-language models through sample-specific test-time optimization~\citep{shu2022tpt,ttrv}. 

Existing test-time reinforcement learning approaches typically sample multiple responses for each test input and derive self-supervision from their agreement, following the self-consistency principle of aggregating diverse reasoning paths~\citep{wang2023selfconsistency}. They commonly construct pseudo-labels through majority voting and assign binary or response-frequency rewards~\citep{ttrl,ttrv}. However, directly applying them to anomalous video understanding poses three critical challenges. (1) Consensus-based pseudo-labels may be unreliable and susceptible to confirmation bias~\citep{wang2026scope}. Because anomalous events are often temporally sparse, visually ambiguous, or dependent on subtle object interactions, a majority answer may not reflect the correct interpretation, and optimizing against it may reinforce existing errors. (2) Conventional binary rewards assign identical rewards to all majority-aligned responses regardless of generation confidence, discarding uncertainty information and weakening the optimization signal~\citep{wang2026scope}. (3) When all responses in a rollout group produce the same answer, identical rewards cause their group-relative advantages to collapse to zero, yielding no effective policy-gradient signals~\citep{shao2024deepseekmath}. Together, these limitations can amplify erroneous pseudo-supervision and weaken test-time optimization.

To address these challenges, we propose a test-time reinforcement learning framework tailored to anomalous video understanding. First, dual-query consistency filtering retains only test samples whose original and semantically reformulated queries yield consistent consensus predictions, improving pseudo-supervision reliability. Second, an entropy-aware consensus reward integrates answer agreement with generation uncertainty to provide a more informative optimization signal than binary majority rewards. Third, a virtual negative anchor introduces reward variation during advantage estimation for unanimous rollout groups, preventing their group-relative advantages from collapsing to zero. The model is optimized on the retained samples using Group Relative Policy Optimization(GRPO), enabling continuous adaptation along the unlabeled test stream.

We evaluate the proposed framework on the three VAU-Bench subsets under
both w/o Think and w/ Think prompting settings. Our method consistently outperforms the frozen Qwen2.5-VL-3B backbone and VAU-R1 in all six QA settings evaluated in this study.
Without additional classification-specific updates or supervision, the QA-adapted models produce consistent improvements in anomaly classification and generally benefit fine-grained classification, indicating cross-task transfer from multiple-choice QA to broader anomaly recognition.

Our main contributions are summarized as follows:

\begin{itemize}
\item To the best of our knowledge, we are the first to introduce online
test-time adaptation into anomalous video understanding, enabling
Video-LLMs to adapt to evolving unlabeled test streams without
ground-truth supervision.

\item We propose an effective test-time reinforcement learning framework
that filters unreliable pseudo-supervision through dual-query
consistency, constructs uncertainty-aware consensus rewards, and
mitigates degenerate group-relative advantages using a virtual negative
anchor.

\item Extensive experiments on VAU-Bench demonstrate consistent
improvements across different prompting settings and validate the
effectiveness of the proposed components and training strategy.
\end{itemize}

%% file: section/related_work.tex
\section{Related Work}

\paragraph{Video Large Language Models.}
Video-LLMs extend large language models
with visual encoders for video-conditioned dialogue and reasoning.
Video-LLaVA aligns image and video representations in a shared language
space, while InternVideo2 scales multimodal pre-training for general
video recognition and video-language tasks
\citep{lin2024videollava,wang2024internvideo2}. TimeChat further
introduces timestamp-aware temporal modeling for long-video
understanding~\citep{ren2024timechat}. Later studies improve video
instruction tuning with large-scale synthetic data and enhance reasoning
through structured reasoning distillation
\citep{zhang2025llavavideo,shi2025aotd}. However, most
Video-LLMs are trained offline and frozen during inference,
limiting their adaptation to rare anomaly patterns and deployment
shifts.

\paragraph{Anomalous Video Understanding.}
Traditional video anomaly detection mainly identifies whether and when
abnormal events occur through weakly supervised multiple-instance
learning or temporal feature modeling
\citep{sultani2018ucfcrime,tian2021rtfm}. Recent methods introduce
vision--language models for stronger semantic generalization. VadCLIP
adapts CLIP for weakly supervised detection, while open-vocabulary and
language-guided open-world methods generalize to unseen anomaly
categories and changing anomaly definitions
\citep{wu2024vadclip,wu2024ovvad,liu2026lagovad}. LAVIDA further uses
pseudo-anomaly exposure and MLLM-based semantic modeling for zero-shot
detection~\citep{dai2026lavida}. Training-free and agentic approaches
include LAVAD, AnomalyRuler, and PANDA, which respectively leverage
LLM-based temporal aggregation, scene-specific rules, and agentic
planning with reflection and memory
\citep{zanella2024lavad,yang2024anomalyruler,yang2025panda}. Beyond
binary detection, Holmes-VAU and VERA support long-term or explainable
anomaly understanding, while VAU-R1 and Vad-R1 enhance anomaly reasoning
through reinforcement fine-tuning and structured Chain-of-Thought
supervision
\citep{holmesvau,ye2025vera,vaur1,huang2025vadr1}. However, these methods
mainly rely on offline training or inference-time prompting and cannot
continuously update model parameters from incoming unlabeled videos to
adapt to evolving scenes, environments, or anomaly types.

\paragraph{Test-time Reinforcement Learning.}
Test-time adaptation addresses deployment shifts by updating model
parameters with unlabeled test inputs. Early methods use self-supervised
objectives or entropy minimization, while later approaches improve
continual adaptation, sample reliability, and resistance to forgetting
\citep{sun2020ttt,wang2021tent,wang2022cotta,niu2022eata}. Test-time
prompt tuning further extends sample-specific adaptation to
vision--language models~\citep{shu2022tpt}. More recently, test-time
reinforcement learning enables large models to learn from self-generated
feedback. TTRL derives pseudo-rewards from agreement among sampled
responses, while TTRV extends this paradigm to vision--language models
\citep{ttrl,ttrv}. SCOPE further improves majority-based optimization
through confidence-weighted pseudo-label estimation
\citep{wang2026scope}. Unlike test-time scaling, these methods update
model parameters and retain acquired knowledge across the test stream.
However, directly applying consensus-based optimization to anomalous
video understanding remains unreliable: sparse and ambiguous evidence
may produce incorrect pseudo-labels, binary rewards ignore generation
uncertainty, and unanimous rollout groups provide no effective
group-relative optimization signal.
\begin{figure*}[t]
    \centering
    \includegraphics[width=\textwidth]
    {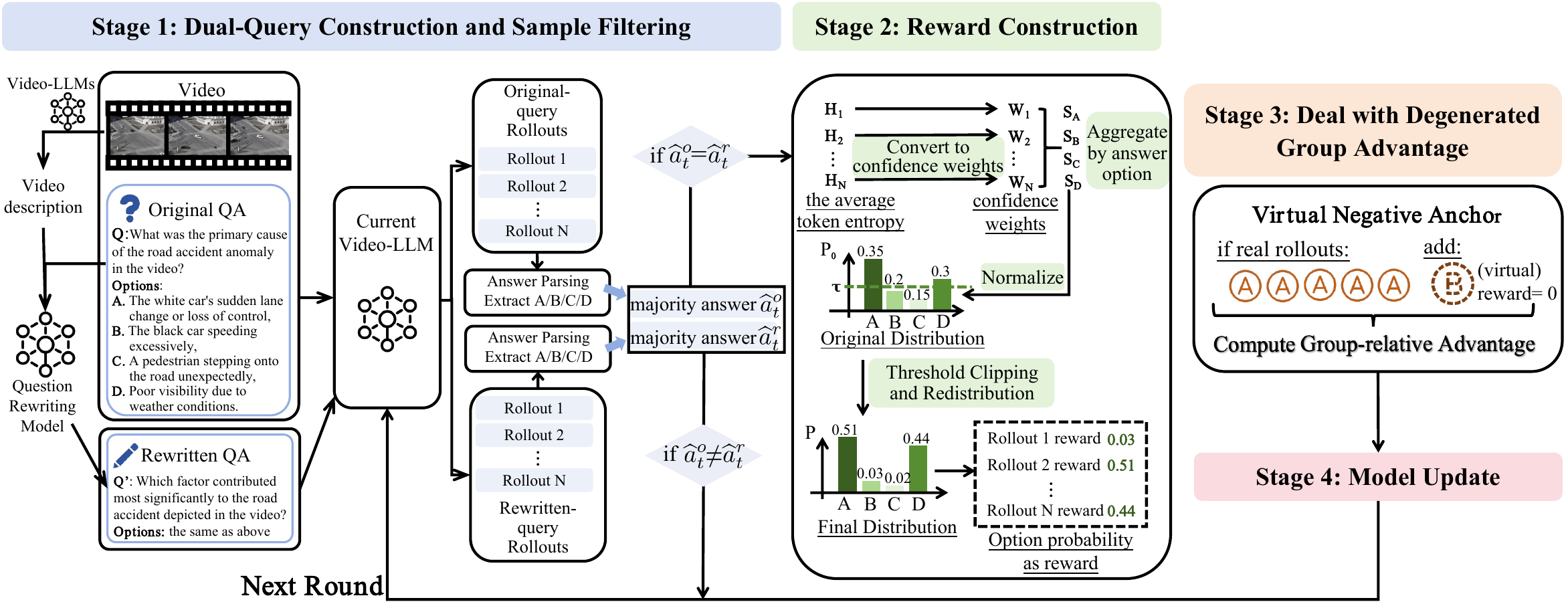}
    \caption{
    Overview of the proposed online test-time reinforcement learning
    framework. Original and reformulated questions are used to identify
    reliable test samples. Retained samples are optimized using an
    entropy-aware consensus reward, and virtual-anchor-enhanced advantage
    estimation if needed.
    }
    \label{fig:method_overview}
\end{figure*}

%% file: section/method.tex
\section{Method}

\subsection{Overview}
We consider test-time reinforcement learning for anomalous video understanding, instantiated as multiple-choice QA following existing benchmarks. At inference, given a sequence of test samples $\{x_t\}_{t=1}^{T}$, let $x_t=(v_t,q_t,\mathcal{O}_t)$ denote the $t$-th sample, where $v_t$, $q_t$, and $\mathcal{O}_t$ represent the input video, original question, and multiple-choice options. The ground-truth answer is unavailable during test-time adaptation. Given the current Video-LLM $\pi_{\theta_t}$ and test sample $x_t$, the goal is to adapt the model on this sample and predict the final option $\widehat{a}_t$. The adapted parameters are carried forward, enabling the model to progressively improve its anomalous video understanding over the test stream. The framework can also be extended to other anomalous video understanding tasks.

To obtain reliable and informative self-supervision while preserving effective optimization on unlabeled test samples, we design the four-stage framework illustrated in Figure~\ref{fig:method_overview}. First, to filter unreliable pseudo-labels, the original question is rewritten into a semantically equivalent form while retaining the original multiple-choice options. The current Video-LLM then samples multiple responses to both question forms, retaining only samples with consistent cross-query consensus. Second, for each retained sample, we construct an entropy-aware consensus reward that jointly considers answer agreement and generation uncertainty. Third, a virtual negative anchor introduces reward variation into unanimous rollout groups, yielding non-zero group-relative advantages and effective optimization signals. Finally, the model is updated using GRPO, performs deterministic inference on the current sample, and is carried forward.

\subsection{Dual-Query Construction and Sample Filtering}

\subsubsection{Semantics-Preserving Query Reformulation}
To assess whether self-generated consensus is robust to linguistic
variation, we construct a semantically equivalent query view for each
test question. A video-LLM first generates a description of the input
video. The video description, original question, and answer options are
then provided to a question-rewriting model, which is instructed to
rewrite only the question while preserving its original meaning. The
answer options and their order remain unchanged.
The original and reformulated inputs are denoted as \(x_t^{o}=(v_t,q_t,\mathcal{O}_t)\) and \(x_t^{r}=(v_t,\widetilde{q}_t,\mathcal{O}_t)\), where \(\widetilde{q}_t\) is the reformulated question.

\subsubsection{Sampling and Consensus-Based Sample Filtering}
To select more reliable samples for test-time optimization, we retain only those with consistent consensus across both query views.
The current policy $\pi_{\theta_t}$
independently generates $N$ stochastic responses for each view:
\(\mathcal{Y}_t^{o}
=
\{y_{t,1}^{o},\ldots,y_{t,N}^{o}\}\)
and
\(\mathcal{Y}_t^{r}
=
\{y_{t,1}^{r},\ldots,y_{t,N}^{r}\}\).
For each view $u\in\{o,r\}$, we extract an answer
label from each response using
\begin{equation}
a_{t,n}^{u}
=
\operatorname{Extract}\!\left(y_{t,n}^{u}\right),
\qquad
{a}_{t,n}^{u}
\in
\mathcal{A}\cup\{\varnothing\},
\end{equation}
where $\mathcal{A}=\{A,B,C,D\}$ is the candidate answer-option set and
$\varnothing$ denotes a response from which no valid option can be
parsed. For each view $u\in\{o,r\}$, the support count for option
$a$ is then computed as
\begin{equation}
c_t^{u}(a)
=
\sum_{n=1}^{N}
\mathbf{1}
\left[
{a}_{t,n}^{u}=a
\right],
\qquad
a\in\mathcal{A}.
\end{equation}

The consensus answer for each view $u\in\{o,r\}$ is
\begin{equation}
\widehat{a}_t^{u}
=
\arg\max_{a}c_t^{u}(a),
\qquad
a\in\mathcal{A}.
\end{equation}

A sample is retained only when both views have a strict majority and
their consensus answers are identical:
\begin{align}
s_t={}&
\mathbf{1}
\left[
\max_a c_t^{o}(a)>\frac{N}{2}
\right]
\mathbf{1}
\left[
\max_a c_t^{r}(a)>\frac{N}{2}
\right]
\nonumber\\
&\cdot
\mathbf{1}
\left[
\widehat{a}_t^{o}=\widehat{a}_t^{r}
\right].
\label{eq:dual_query_filter}
\end{align}

For example, when $N=4$, suppose the extracted answers for the original
and reformulated queries are $(A,A,A,B)$ and $(A,A,B,C)$, respectively.
Although both views yield the same consensus answer,
$\widehat{a}_t^{o}=\widehat{a}_t^{r}=A$, their support counts are
$c_t^{o}(A)=3$ and $c_t^{r}(A)=2$. Since the reformulated query does not
satisfy the strict-majority condition
$c_t^{r}(A)>N/2$, the sample is discarded with $s_t=0$.

Samples with $s_t=0$ do not contribute to parameter update. For
samples with $s_t=1$, only the original-question rollouts are used for
reward construction, while the reformulated-question
rollouts are used solely for reliability assessment.

\begin{figure}[t]
\centering
\includegraphics[width=\columnwidth]{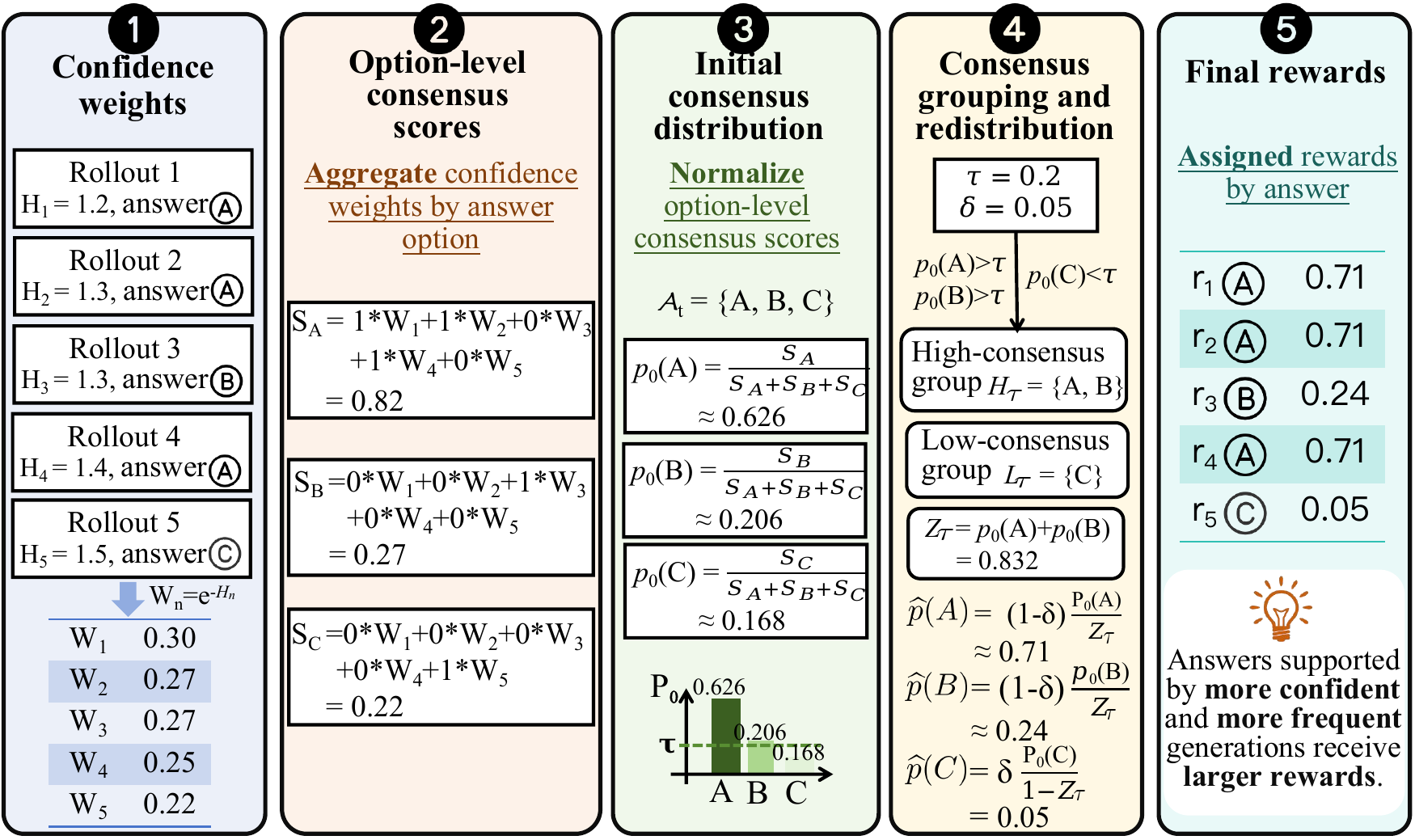}
    \caption{An example of the entropy-aware consensus reward
    computation with five rollouts retained from the first stage.}
    \label{fig:entropy_reward_example}
\end{figure}

\subsection{Entropy-Aware Consensus Reward}

To provide more informative supervision than a binary majority reward,
we construct an entropy-aware consensus reward for each retained test
sample. We use only the rollouts generated from the original query for
reward estimation and policy optimization. Figure~\ref{fig:entropy_reward_example} provides an example of
the complete entropy-aware consensus reward computation process.

Let $L_n$ denote the number of generated tokens in the $n$-th
original-query rollout, and let $P_{n,\ell}(v)$ be the probability of
vocabulary token $v$ at generation position $\ell$. The mean token
entropy of the rollout is
\begin{equation}
H_n
=
-\frac{1}{L_n}
\sum_{\ell=1}^{L_n}
\sum_{v\in\mathcal{V}}
P_{n,\ell}(v)\log P_{n,\ell}(v),
\label{eq:entropy}
\end{equation}
and its corresponding confidence weight is
\begin{equation}
w_n
=
\exp(- H_n),
\label{eq:weight}
\end{equation}
A rollout with lower mean token entropy therefore receives
a larger confidence weight.

To obtain an option-level consensus score, we
first group the rollouts according to their extracted answer labels and
then sum the confidence weights within each group. Specifically, for
each answer label $a\in\mathcal{A}$,
\begin{equation}
S_t(a)
=
\sum_{n=1}^{N}
\mathbf{1}
\left[
a_{t,n}^{o}=a
\right]w_n
\label{eq:consensus_score}
\end{equation}
Thus, $S_t(a)$ measures the confidence-weighted support received by
option $a$: it increases when more rollouts select $a$ or when those
rollouts are generated with lower uncertainty.

We denote $\mathcal{A}_t $ as 
the set of answer labels that actually appear among the $N$
original-query rollouts. The option-level scores are normalized into an initial
consensus distribution:
\begin{equation}
p_0(a\mid x_t^{o})
=
\frac{
S_t(a)
}{
\sum_{b\in\mathcal{A}_t}S_t(b)
},
\qquad
a\in\mathcal{A}_t.
\label{eq:initial_distribution}
\end{equation}

To prevent weakly supported answers from introducing noisy optimization signals, we apply a consensus threshold to divide the observed answer labels into high- and low-consensus groups:
\begin{equation}
\mathcal{H}_{\tau,t}
=
\left\{
a\in\mathcal{A}_t
:
p_0(a\mid x_t^{o})\geq\tau
\right\},
\qquad
\mathcal{L}_{\tau,t}
=
\mathcal{A}_t\setminus\mathcal{H}_{\tau,t},
\label{eq:consensus_groups}
\end{equation}
where $\tau$ is the consensus threshold. The total probability assigned
to the high-consensus group is
\begin{equation}
Z_{\tau,t}
=
\sum_{a\in\mathcal{H}_{\tau,t}}
p_0(a\mid x_t^{o}).
\label{eq:high_consensus_mass}
\end{equation}

Rather than assigning zero reward to low-consensus answers, we retain a small probability mass to avoid overly hard supervision and preserve plausible alternatives:
\begin{equation}
\widehat{p}(a\mid x_t^{o})
=
\begin{cases}
\displaystyle
(1-\delta)
\frac{p_0(a\mid x_t^{o})}{Z_{\tau,t}},
&
a\in\mathcal{H}_{\tau,t},
\\[9pt]
\displaystyle
\delta
\frac{p_0(a\mid x_t^{o})}{1-Z_{\tau,t}},
&
a\in\mathcal{L}_{\tau,t}.
\end{cases}
\label{eq:redistributed_distribution}
\end{equation}
Here, $\tau$ determines which answer labels receive high-consensus
status, whereas $\delta$ controls the small probability mass retained
for the remaining observed labels. If no low-consensus label exists, we retain the initial distribution $p_0$.

Finally, each rollout receives the probability assigned to its
extracted answer, or zero if no valid answer is parsed:
\begin{equation}
r_n =
\begin{cases}
\widehat{p}(a_{t,n}^{o}\mid x_t^{o}), & a_{t,n}^{o}\in\mathcal{A}_t,\\
0, & a_{t,n}^{o}=\varnothing.
\end{cases}
\label{eq:rollout_reward}
\end{equation}
Thus, rollouts supporting the same answer receive the same reward,
while unparseable responses receive no positive reinforcement.

\subsection{Virtual-Anchor Advantage Estimation}

GRPO computes group-relative advantages by normalizing the rollout
rewards:
\begin{equation}
A_n = 
\frac{r_n-\mu_{\mathcal{R}}}
{\sigma_{\mathcal{R}}+\epsilon},
\label{eq:group_advantage}
\end{equation}
where $\mathcal{R}={r_1,\ldots,r_N}$, and
$\mu_{\mathcal{R}}$ and $\sigma_{\mathcal{R}}$ denote its mean and
standard deviation, respectively.

When all rollouts are parsed into the same answer and receive identical
rewards, $\sigma_{\mathcal{R}}=0$, causing their group-relative
advantages to vanish and eliminating effective policy-gradient signals.
To prevent unanimous rollout groups from yielding zero group-relative
advantages and ineffective gradients, inspired by virtual-sample
augmentation~\citep{he2026advantagecollapse}, we introduce a virtual negative anchor and temporarily augment the reward set as
$\widetilde{\mathcal{R}}
={r_1,\ldots,r_N,r_{\mathrm{va}}}$, where
$r_{\mathrm{va}}=0$ is a virtual reward.
The mean and standard deviation in
Equation~\eqref{eq:group_advantage} are then computed over
$\widetilde{\mathcal{R}}$ rather than $\mathcal{R}$, with
$\epsilon=10^{-12}$. Only the advantages of the $N$ real rollouts are
retained. The virtual negative anchor affects optimization solely through these normalization statistics and is excluded from the GRPO loss computation.

\subsection{Optimization}
Finally, the model parameters are updated by minimizing the standard
GRPO loss using the resulting advantages, with a KL penalty. The adapted model then performs
deterministic inference on the current sample, and its parameters are
carried forward to the next test sample, forming a continuous online
adaptation process.

%% file: section/experiment.tex
\section{Experiments}

\subsection{Experimental Setup}

\paragraph{Dataset and metrics.}
We evaluate on the test split of VAU-Bench \citep{vaur1}, which
integrates MSAD, UCF-Crime, and ECVA and provides annotations for
multiple-choice QA and anomaly classification over 19 major anomaly
categories. We exclude 28 corrupted ECVA videos from the original 929
samples, resulting in 901 valid videos: 240 from MSAD, 251 from
UCF-Crime, and 410 from ECVA. Only these unlabeled test samples are used
for test-time training and evaluation; ground-truth labels are
accessed solely for metric computation. We report multiple-choice
accuracy for QA and binary and multi-class accuracy for anomaly
classification. Following VAU-R1, results are reported under
\emph{w/o Think} and \emph{w/ Think} prompting settings.

\paragraph{Baselines.}
We compare with the frozen Qwen2.5-VL-3B-Instruct backbone and VAU-R1
\citep{vaur1}. VAU-R1 uses the same backbone and is trained on the
VAU-Bench training split through supervised reinforcement learning
across multiple anomaly-understanding tasks. All methods follow the
same prompting and evaluation protocols.

\paragraph{Implementation details.}
We use Qwen2.5-VL-3B-Instruct as the trainable policy backbone for test-time adaptation, while separate frozen instances of the same model are employed for video description generation and question reformulation.We perform full-parameter adaptation with a learning
rate of $1\times10^{-6}$, a batch size of 2, and one update per batch.
GRPO samples $N=3$ rollouts with temperature 1.0 and top-$p$ 1.0. We
set $\beta=0.04$, $\tau=0.2$, and $\delta=0.05$. The maximum completion
length is 20 tokens for w/o Think and 512 tokens for w/ Think.
For QA, videos are sampled at 1 FPS with at most 60 frames; classification
uses 1 FPS without a frame limit. Each subset is processed independently
from the original backbone, with the test order shuffled.
For every batch, the model first adapts on the unlabeled inputs and then
performs deterministic inference on the same batch, while updated
parameters are retained for subsequent batches. Samples rejected by
dual-query filtering are excluded from parameter updates. For cross-task
evaluation, each QA-adapted model is directly evaluated on the binary
and multi-class classification tasks of the corresponding subset,
without further updates or supervision. Experiments are
conducted on NVIDIA A800 and H100 GPUs.

\subsection{Main Results}

\begin{table}[!t]
    \centering
    \small
    \setlength{\tabcolsep}{3.5pt}
    \renewcommand{\arraystretch}{1.08}
    \resizebox{0.45\textwidth}{!}{%
    \begin{tabular}{l l c c}
        \hline
        \textbf{Dataset}
        & \textbf{Model}
        & \textbf{w/o Think}
        & \textbf{w/ Think} \\
        \hline

        \multirow{3}{*}{MSAD}
        & Qwen2.5-VL-3B & 85.83 & 82.50 \\
        & VAU-R1        & 88.33 & 87.08 \\
        & \cellcolor{oursblue} Ours
        & \cellcolor{oursblue} \textbf{92.50} {\scriptsize(+6.67)}
        & \cellcolor{oursblue} \textbf{90.00} {\scriptsize(+7.50)} \\
        \hline

        \multirow{3}{*}{UCF-Crime}
        & Qwen2.5-VL-3B & 91.63 & 83.27 \\
        & VAU-R1        & 92.03 & 91.63 \\
        & \cellcolor{oursblue} Ours
        & \cellcolor{oursblue} \textbf{92.83} {\scriptsize(+1.20)}
        & \cellcolor{oursblue} \textbf{92.43} {\scriptsize(+9.16)} \\
        \hline

        \multirow{3}{*}{ECVA}
        & Qwen2.5-VL-3B & 85.58 & 75.81 \\
        & VAU-R1        & 89.53 & 86.51 \\
        & \cellcolor{oursblue} Ours
        & \cellcolor{oursblue} \textbf{91.46} {\scriptsize(+5.88)}
        & \cellcolor{oursblue} \textbf{90.00} {\scriptsize(+14.19)} \\
        \hline
    \end{tabular}
    }
    \caption{
        Multiple-choice QA accuracy (\%) on VAU-Bench.
        Best results are highlighted in bold.
    }
    \label{tab:qa_main}
\end{table}

\begin{table}[!t]
    \centering
    \small
    \setlength{\tabcolsep}{3.5pt}
    \renewcommand{\arraystretch}{1.08}
    \resizebox{0.48\textwidth}{!}{%
    \begin{tabular}{l l c c}
        \hline
        \multirow{2}{*}{\textbf{Dataset}}
        & \multirow{2}{*}{\textbf{Model}}
        & \textbf{w/o Think}
        & \textbf{w/ Think} \\
        \cline{3-4}
        &
        & \textbf{Bin. Acc./Multi. Acc.}
        & \textbf{Bin. Acc./Multi. Acc.} \\
        \hline

        \multirow{3}{*}{MSAD}
        & Qwen2.5-VL-3B
        & 79.17 / 69.58
        & 73.33 / 56.67 \\
        & VAU-R1
        & 82.08 / 71.25
        & 74.58 / 60.83 \\
        & \cellcolor{oursblue}Ours
        & \cellcolor{oursblue}\textbf{87.50} / \textbf{73.75}
        & \cellcolor{oursblue}\textbf{78.33} / \textbf{62.61} \\
        \hline

        \multirow{3}{*}{UCF-Crime}
        & Qwen2.5-VL-3B
        & 64.54 / 58.57
        & 62.55 / 52.19 \\
        & VAU-R1
        & 62.55 / 57.77
        & 62.15 / \textbf{56.97} \\
        & \cellcolor{oursblue}Ours
        & \cellcolor{oursblue}\textbf{78.88} / \textbf{62.55}
        & \cellcolor{oursblue}\textbf{67.33} / 54.58 \\
        \hline

        \multirow{3}{*}{ECVA}
        & Qwen2.5-VL-3B
        & 52.83 / 30.16
        & 49.89 / 22.00 \\
        & VAU-R1
        & 49.66 / 30.61
        & 55.78 / \textbf{31.07} \\
        & \cellcolor{oursblue}Ours
        & \cellcolor{oursblue}\textbf{64.88} / \textbf{33.90}
        & \cellcolor{oursblue}\textbf{56.83} / 25.85 \\
        \hline
    \end{tabular}
    }
    \caption{
Binary and multi-class anomaly classification accuracy (\%).
Bin.Acc. and Multi.Acc. denote binary and multi-class accuracy, respectively.
}
    \label{tab:classification}
\end{table}

\subsubsection{Performance on Multiple-Choice QA Task}  Table~\ref{tab:qa_main} reports the multiple-choice QA results on the
three VAU-Bench subsets. Our method consistently outperforms the frozen Qwen2.5-VL-3B backbone and VAU-R1 in all six QA settings evaluated in this study. Compared with the frozen
Qwen2.5-VL-3B backbone, it improves accuracy by 6.67, 1.20, and 5.88 percentage points under the w/o Think setting, and by 7.50, 9.16, and
14.19 percentage points under the w/ Think setting on MSAD, UCF-Crime, and ECVA,
respectively. It also consistently outperforms VAU-R1, indicating that
offline supervised training does not fully eliminate the need to adapt
to the incoming test distribution.
The largest gain is observed on ECVA with explicit reasoning, where
accuracy increases from 75.81\% to 90.00\%. In contrast, the smaller
improvement on UCF-Crime without reasoning is partly associated with its
already strong frozen baseline of 91.63\%, which leaves limited room for
further gains. Overall, the consistent improvements across different
datasets support the effectiveness of online adaptation using unlabeled
test samples. The larger gains under the w/ Think setting further
suggest that reasoning-based generations provide useful signals for
self-supervised test-time optimization.

\subsubsection{Performance on Anomaly Classification Task}
Table~\ref{tab:classification} evaluates whether the knowledge acquired
through multiple-choice QA adaptation transfers to anomaly
classification. Without any additional parameter updates or
classification-label supervision, our method consistently outperforms
the frozen Qwen2.5-VL-3B backbone across all datasets, prompting
settings, and classification tasks. For binary classification, the
improvements range from 4.78 to 14.34 percentage points, with the
largest gain observed on UCF-Crime under the w/o Think setting. For
fine-grained multi-class classification, our method also improves the
frozen backbone in all six settings, demonstrating that QA-based
test-time adaptation benefits not only anomaly detection but also
category-level recognition.
These results provide evidence that adaptation driven solely by
multiple-choice QA responses can enhance general anomaly-related
representations and transfer across tasks. VAU-R1 is additionally
reported as a supervised reference model, since it has been jointly
trained on multiple labeled tasks, including anomaly classification.
Despite using no classification supervision during adaptation, our
method remains competitive with, and in most settings surpasses,
VAU-R1, further supporting the effectiveness of the learned
test-time representations.

\subsection{Insightful Analysis}

\subsubsection{Ablation of Reward Designs}

Table~\ref{tab:reward_comparison} compares our entropy-aware consensus
reward with a TTRL-style binary reward under the same adaptation pipeline. Although the
binary reward consistently improves over the frozen Video-LLM, our
reward achieves higher accuracy in all six settings. Specifically, it
outperforms the binary reward by 2.92, 0.80, and 0.97 percentage points under w/o Think, and by 1.25, 2.39, and 2.20 percentage points under w/ Think on
MSAD, UCF-Crime, and ECVA, respectively. The larger average gain under
w/ Think suggests that jointly considering answer agreement and
generation confidence provides more informative supervision for longer responses.

\begin{table}[t]
    \centering
    \small
    \setlength{\tabcolsep}{3.5pt}
    \renewcommand{\arraystretch}{1.05}
    \begin{tabular}{l l c c}
        \hline
        \textbf{Dataset}
        & \textbf{Reward}
        & \textbf{w/o Think}
        & \textbf{w/ Think} \\
        \hline

        \multirow{3}{*}{MSAD}
        & Qwen2.5-VL-3B  & 85.83 & 82.50 \\
        & Binary $0/1$ reward & 89.58 & 88.75 \\
        & Entropy-Aware Reward & \textbf{92.50} & \textbf{90.00} \\
        \hline

        \multirow{3}{*}{UCF-Crime}
        & Qwen2.5-VL-3B & 91.63 & 83.27 \\
        & Binary $0/1$ reward & 92.03 & 90.04 \\
        & Entropy-Aware Reward & \textbf{92.83} & \textbf{92.43} \\
        \hline

        \multirow{3}{*}{ECVA}
        & Qwen2.5-VL-3B & 85.58 & 75.81 \\
        & Binary $0/1$ reward & 90.49 & 87.80 \\
        & Entropy-Aware Reward & \textbf{91.46} & \textbf{90.00} \\
        \hline
    \end{tabular}
    \caption{
        Controlled comparison of reward designs. 
    }
    \label{tab:reward_comparison}
\end{table}

\subsubsection{Component Ablation}

\begin{table}[!t]
    \centering
    \small
    \setlength{\tabcolsep}{3.5pt}
    \renewcommand{\arraystretch}{1.08}
    \begin{tabular}{l c c c}
        \hline
        \multirow{2}{*}{\textbf{Configuration}}
        & \textbf{MSAD}
        & \textbf{UCF-Crime}
        & \textbf{ECVA} \\
        \cline{2-4}
        & \textbf{w/o T / w/ T}
        & \textbf{w/o T / w/ T}
        & \textbf{w/o T / w/ T} \\
        \hline

        Qwen2.5-VL-3B
        & 85.83 / 82.50
        & 91.63 / 83.27
        & 85.58 / 75.81 \\

        + ER
        & 91.67 / 89.17
        & 92.03 / 90.44
        & 88.54 / 88.54 \\

        + ER + VA
        & 92.08 / 89.58
        & 92.43 / 90.84
        & 90.98 / 88.78 \\

        + ER + VA + DQ
        & \textbf{92.50} / \textbf{90.00}
        & \textbf{92.83} / \textbf{92.43}
        & \textbf{91.46} / \textbf{90.00} \\
        \hline
    \end{tabular}
    \caption{
        Cumulative component analysis. Here, w/o T and w/ T abbreviate w/o Think and w/ Think, respectively. ER, VA, and DQ denote the entropy-aware reward, virtual negative anchor, and dual-query filtering.}
    \label{tab:component_ablation}
\end{table}

\begin{table}[!t]
\centering
\small

\begin{subtable}[t]{0.46\columnwidth}
    \centering
    \setlength{\tabcolsep}{8pt}
    \renewcommand{\arraystretch}{1.05}
    \begin{tabular}{c c}
        \toprule
        \textbf{Batch Size}
        & \textbf{Acc.} \\
        \midrule
        1 & \textbf{92.92} \\
        2 & 92.50 \\
        4 & 90.42 \\
        8 & 88.75 \\
        \bottomrule
    \end{tabular}
    \caption{Effect of batch size.}
    \label{tab:batch_size}
\end{subtable}
\hfill
\begin{subtable}[t]{0.46\columnwidth}
    \centering
    \setlength{\tabcolsep}{8pt}
    \renewcommand{\arraystretch}{1.05}
    \begin{tabular}{c c}
        \toprule
        \textbf{Numbers}
        & \textbf{Acc.} \\
        \midrule
        2 & \textbf{93.75} \\
        3 & 92.50 \\
        4 & 89.17 \\
        \bottomrule
    \end{tabular}
    \caption{Effect of sampled numbers.}
    \label{tab:numbers}
\end{subtable}

\caption{
    Hyperparameter analysis of batch size and sampled numbers on MSAD subset of VAU-Bench under the w/o Think setting.
}
\label{tab:hyperparameters}

\end{table}

Table~\ref{tab:component_ablation} presents a cumulative ablation of
the three proposed components. Introducing the entropy-aware consensus
reward yields the largest initial improvement over the frozen VLM,
increasing accuracy across all datasets and prompting settings. The
gains are particularly pronounced under the w/ Think setting, where
longer generations contain richer uncertainty information that can be
exploited during reward construction. Adding the virtual negative
anchor mechanism further improves all six settings, confirming that
restoring reward variation for unanimous rollout groups provides more
effective group-relative optimization signals.
Dual-query consistency filtering brings additional improvements in
every setting, with relatively larger gains on ECVA and UCF-Crime under
w/ Think. This result supports its role in reducing the influence of
unreliable pseudo-supervision before model updating. Overall, the full
framework improves the frozen backbone by 6.67/7.50, 1.20/9.16, and
5.88/14.19 percentage points on MSAD, UCF-Crime, and ECVA under the
w/o Think/w/ Think settings, respectively. The consistently increasing
performance across the cumulative configurations demonstrates that the
three components are complementary and each contributes to the final
adaptation performance.

\subsubsection{Hyperparameter Analysis}

\begin{figure*}[!t]
    \centering
    \includegraphics[width=\textwidth]
    {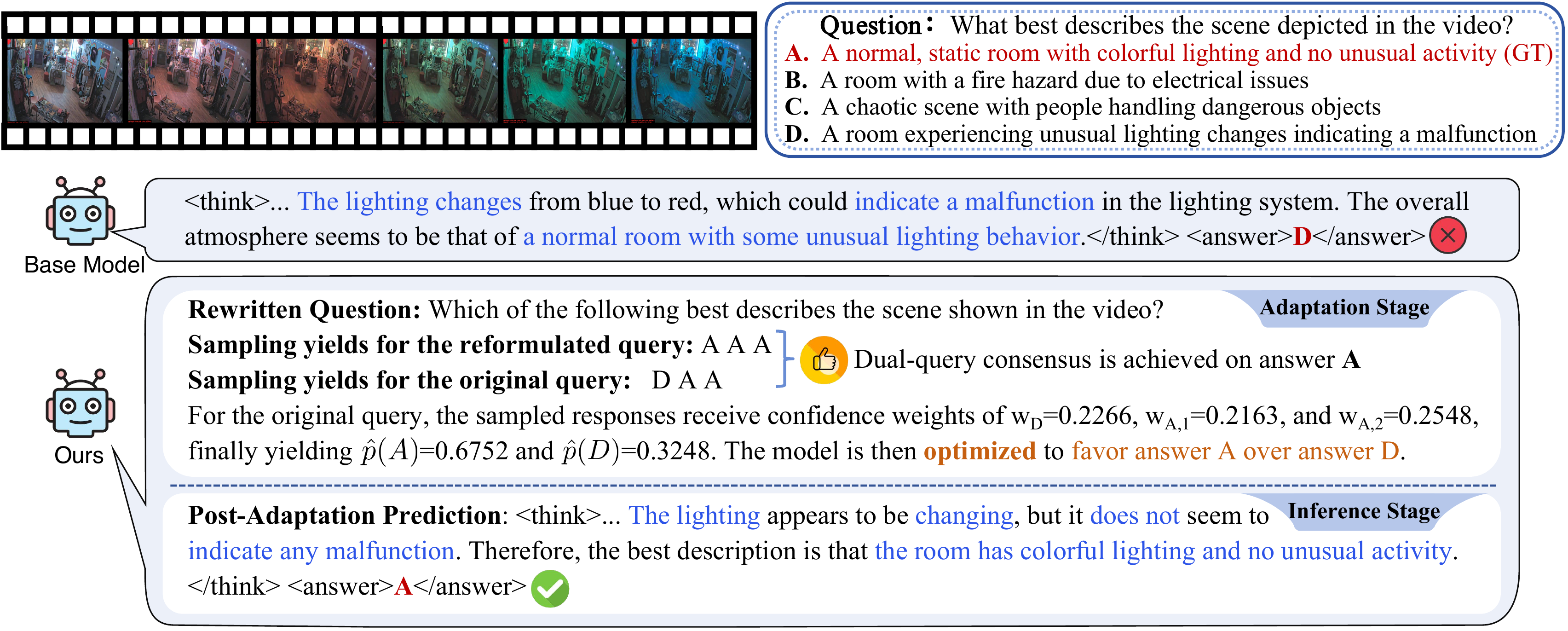}
    \caption{Qualitative case study of multiple-choice QA. The base model Qwen2.5-VL-3B misinterprets
lighting changes as a malfunction, whereas our method uses
dual-query consistency and entropy-aware rewards to correct the prediction
from D to A.}
    \label{fig:case_study}
\end{figure*}

Table~\ref{tab:hyperparameters} investigates the effects of batch size and sampled generations $N$ on MSAD under the w/o Think setting. For the batch-size study, the number of generations is fixed at 3, whereas for the generation-number study, the batch size is fixed at 2. As shown in Table~\ref{tab:batch_size}, a batch size of 1 achieves the highest accuracy, while a batch size of 2 yields a comparable result. Increasing the batch size to 4 or 8 causes performance degradation. This suggests that smaller batches are more suitable for test-time adaptation, as they allow more frequent updates and reduce interference from jointly optimizing heterogeneous test samples.
Regarding the number of generations, Table~\ref{tab:numbers} shows that $N=2$ achieves the highest accuracy, whereas further increasing $N$ provides no additional gains. Larger rollout groups may introduce more low-quality or uncertain responses, thereby weakening the consensus signal. Nevertheless, we use $N=3$ in the main experiments to enable
non-unanimous majority voting at a moderate sampling cost. By contrast, under the strict-majority criterion, $N=2$ retains only unanimous groups. Overall, these results indicate that test-time adaptation benefits from compact batches and rollout groups rather than simply increasing either quantity.

\subsubsection{Error Analysis}
\begin{figure}[t]
    \centering
    \includegraphics[width=0.98\linewidth]{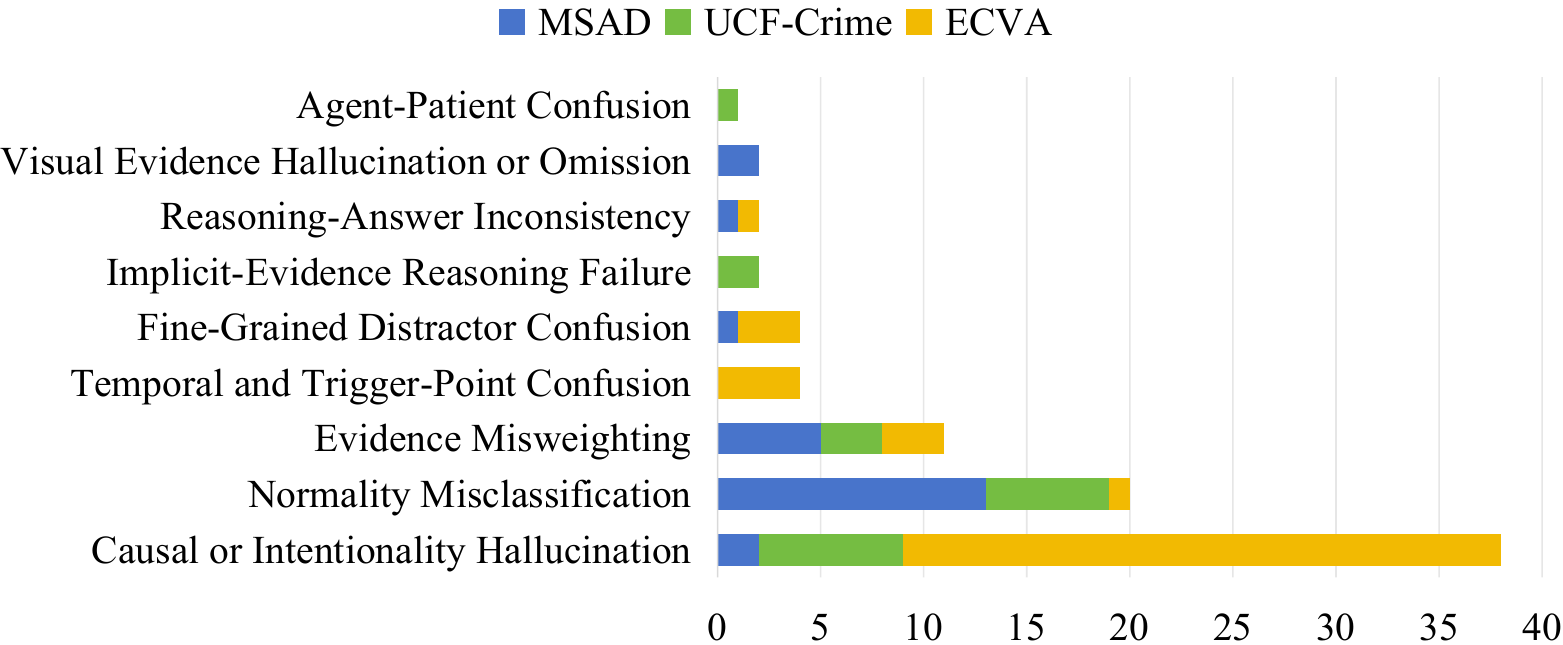}
    \caption{Distribution of model failure modes across MSAD, UCF-Crime, and
    ECVA. The bars report the number of errors attributed to each
    failure mode.}
    \label{fig:error_distribution}
    \vspace{-6mm}
\end{figure}
We exhaustively categorize the 84 incorrect predictions produced under
the w/ Think setting, including 41
from ECVA, 24 from MSAD, and 19 from UCF-Crime.
Causal or intentionality hallucination is the dominant failure mode
(38/84, 45.24\%), indicating that the model frequently substitutes
unsupported commonsense explanations for observable evidence.
Normality misclassification ranks second (20/84, 23.81\%), especially on
MSAD and UCF-Crime, revealing an overly strong anomaly prior and
insufficient use of negative evidence. Evidence misweighting accounts
for another 11 errors (13.10\%), where relevant cues are detected but the
most discriminative evidence is not selected. The remaining errors
mainly involve temporal reasoning, fine-grained distractor confusion,
and reasoning--answer inconsistency. These results suggest that the
primary bottleneck lies in evidence-grounded causal reasoning, normality
calibration, and option-level evidence comparison rather than basic
anomaly recognition.

\subsection{Case Study}
Figure~\ref{fig:case_study} illustrates how our method corrects an
error caused by unreliable evidence prioritization. The base model incorrectly selects option D by over-interpreting
the changing illumination as evidence of a malfunction, despite the
absence of explicit anomalous activity. In contrast, stochastic
sampling yields $(D,A,A)$ for the original query and $(A,A,A)$ for its
semantically equivalent reformulation. The consistent cross-query
consensus on A indicates that the erroneous prediction D is unstable,
whereas A reflects a more robust interpretation. Our entropy-aware
reward further assigns a consensus probability of $0.6752$ to A and
$0.3248$ to D, producing positive relative advantages for the
A-supporting rollouts and a negative advantage for D. After adaptation,
the model correctly recognizes that the lighting changes alone do not
indicate a malfunction and predicts A. This case illustrates how our
method recovers latent correct knowledge from stochastic generations and converts it into an
effective self-supervised optimization signal.